\documentclass[conference]{IEEEtran}
\IEEEoverridecommandlockouts
\usepackage[utf8]{inputenc}
\usepackage{cite}
\usepackage{amsmath,amssymb,amsfonts}
\usepackage{algorithmic}
\usepackage{graphicx}
\usepackage{textcomp}
\usepackage{url}
\usepackage{subcaption}
\usepackage[acronym]{glossaries}
\usepackage[normalem]{ulem}
\usepackage{amsfonts}
\usepackage{booktabs}
\usepackage{multirow}
\usepackage[table]{xcolor}
\usepackage{array}
\usepackage{bytefield}
\usepackage{float}
\newacronym{dvs}{DVS}{Dynamic Vision Sensor}
\newacronym{cnn}{CNN}{Convolutional Neural Network}
\newacronym{bnn}{BNN}{Binary Neural Network}
\newacronym{tnn}{TNN}{Ternary Neural Network}
\newacronym{mac}{MAC}{Multiplication-Accumulation}
\newacronym{fpga}{FPGA}{Field-Programmable Gate Array}
\newacronym{pt}{PT}{Pan-Tilt}
\newacronym{relu}{ReLU}{Rectified Linear Unit}
\newacronym{asic}{ASIC}{Application Specific Integrated Circuit}
\newacronym{gpu}{GPU}{Graphic Processing Unit}
\newacronym{tpu}{TPU}{Tensor Processing Unit}
\newacronym{dsp}{DSP}{Digital Signal Processing}
\newacronym{hls}{HLS}{High-level Synthesis}
\newacronym{pol}{PoL}{Point of Load}
\newacronym{adc}{ADC}{analog-to-digital conversion}
\newacronym{ppf}{PPF}{point-process filter}
\newacronym{iot}{IoT}{Internet of Things}
\newacronym{fog}{FOG}{Fog Computing}
\newacronym{pca}{PCA}{principal component analysis}
\newacronym{omd}{OMD}{Object Motion Detector}
\newacronym{fifo}{FIFO}{First-In, First-Out}
\newacronym{lpwa}{LPWA}{Low-power wide-area}
\newacronym{cis}{CIS}{Conventional Image-based Sensors}
\newacronym{dma}{DMA}{Direct Memory Access}
\newacronym{simd}{SIMD}{single instruction, multiple data}
\newacronym{fpu}{FPU}{floating-point unit}
\newacronym{gaer}{G-AER}{group address-event representation}

\ifthenelse{0>1}{%
  \newcommand{\fixme}[1]{{\color{red} FIXME $\Rightarrow$}~{\color{orange}#1}}
  \newcommand{\checkthis}[1]{{\color{red}{#1}}}
  \newcommand{\dansaid}[1]{{\color{brown} DAN SAID:}~{\color{violet}#1}}
  \newcommand{\volkansaid}[1]{{\color{brown} VOLKAN SAID:}~{\color{violet}#1}}
  \newcommand{\paperidea}[1]{{\color{blue} Paper idea:}~{\color{olive}#1}}
  \newcommand{\areweauthorized}[1]{{\color{blue} Are we authorized?}~{\color{olive}#1}}
}{%
  \newcommand{\fixme}[1]{}
  \newcommand{\checkthis}[1]{#1}
  \newcommand{\dansaid}[1]{}
  \newcommand{\volkansaid}[1]{}
  \newcommand{\paperidea}[1]{}
  \newcommand{\areweauthorized}[1]{}
}

\def\BibTeX{{\rm B\kern-.05em{\sc i\kern-.025em b}\kern-.08em
    T\kern-.1667em\lower.7ex\hbox{E}\kern-.125emX}}

\newcommand{\framesizebits}{\checkthis{$1397\,\text{bits}$}}
\newcommand{\percentagesavings}{\checkthis{$99.6\,\%$}}
\newcommand{\accuracyF}{\checkthis{$83\,\%$}}
\newcommand{\energyconsumption}{\checkthis{$91.1\,\text{mW}$}}
\newcommand{\bandwidthtenfps}{\checkthis{$2.23\,\text{kbps}$}}
\newcommand{\bandwidthtenfpstop}{\checkthis{$13.97\,\text{kbps}$}}
\newcommand{\bandwidthfull}{\checkthis{$74.58\,\text{kbps}$}}
\newcommand{\inferencetime}{\checkthis{$450\,\text{ms}$}}
\newcommand{\windowtime}{\checkthis{$3\,\text{ms}$}}
\newcommand{\eventrate}{\checkthis{$50\,\text{M events/s}$}}

\begin{document}

\title{Near-chip Dynamic Vision Filtering for Low-Bandwidth Pedestrian Detection\\
\thanks{*Both authors contributed equally to this work.}
}

\author{\IEEEauthorblockN{Anthony Bisulco*, Fernando Cladera Ojeda*, Volkan Isler, Daniel D. Lee}
\IEEEauthorblockA{\textit{Samsung AI Center NY} \\
New York, NY, USA \\
saic-ny@samsung.com}

}

\maketitle

\begin{abstract}

This paper presents a novel end-to-end system for pedestrian detection
using \glspl{dvs}. We target applications where
multiple sensors transmit data to a local processing unit, which
executes a detection algorithm.
Our system is composed of (i) a
near-chip event filter that compresses and
denoises the event stream from the DVS, and (ii) a \gls{bnn} detection
module that runs on a low-computation edge computing device (in our
case a STM32F4 microcontroller).

We present the system architecture and provide an end-to-end
implementation for pedestrian detection in an office environment.  Our
implementation reduces transmission size by  up to \percentagesavings{}
compared to
transmitting the raw event stream. The
average packet size in our system is only \framesizebits{}, while
$307.2\,\text{kb}$ are required to send an uncompressed \gls{dvs} time
window.
Our detector is able to perform a detection every \inferencetime{}, with
an overall testing F1 score of \accuracyF{}.  The low bandwidth and energy properties
of our system make it ideal for IoT applications.

\end{abstract}


\section{Introduction}
\glsreset{dvs}
\gls{dvs} technologies hold the potential to revolutionize imaging systems by
enabling asynchronous, event-based image acquisition. \gls{dvs} pixels generate and transmit
events only when there is a change in light intensity of a
pixel. This approach has many
advantages compared to \checkthis{\gls{cis}}, such as: (i)~higher dynamic range,
(ii)~higher sampling rates, (iii)~lower bandwidth requirements between the
sensor and the processing unit, and (iv)~lower power consumption. These
characteristics make \glspl{dvs} attractive sensors for energy-constrained
scenarios such as the
\gls{iot} applications.

In this paper, we focus on the application of \gls{dvs} based systems to
pedestrian detection.
A common solution to this problem
involves streaming data from a \gls{cis} to a processing module that runs the detection algorithm. Since
the raw data from the imaging sensor can be overwhelming, usually the
images are compressed before transmission. This approach (i)~requires a large bandwidth or
low frame rate to stream the data in a bandwidth constrained environment, and (ii)~raises inherent privacy concerns, as streamed images
may be accessed by malicious third-party actors.
Inference at the edge~\cite{edgeFacebook},
where data acquisition and processing are performed on-device, has been proposed as a solution for these
problems.
Unfortunately, the amount of energy required for inference at the edge when using
\gls{cis} limits its applicability.
Near-chip feature extraction and data compression has the potential to
provide a middle-ground
solution.

\begin{figure}
  \centering
  \includegraphics[width=1.0\linewidth]{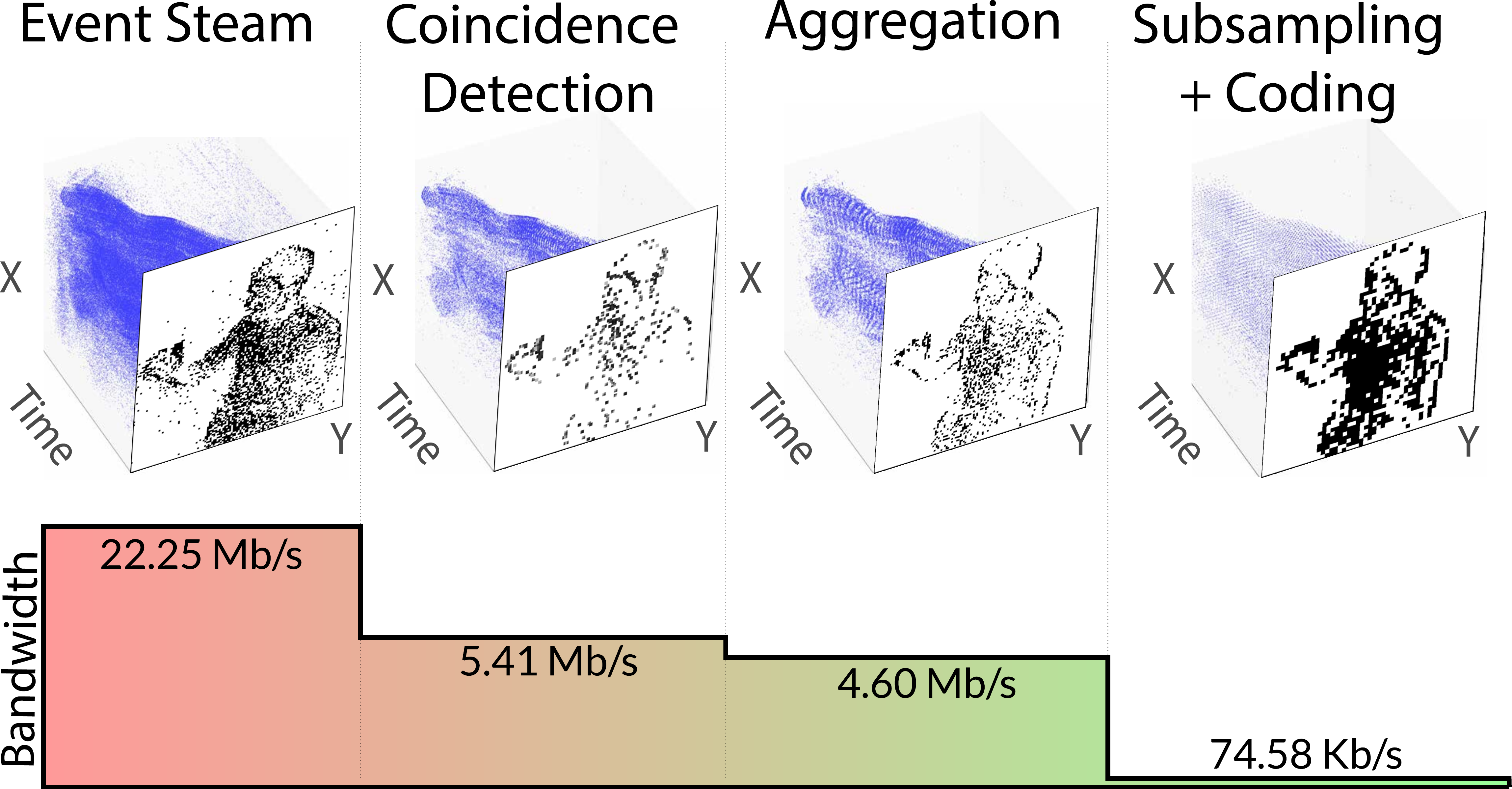}
  \caption{Near-chip \gls{dvs} filter architecture stages. We observe a
    reduction of the required bandwidth across the different stages of the filter,
    as well as a sparser event stream. The privacy is enhanced due to
    lossy  subsampling, as shown by windowed events on the figure. }
  \label{fig:bwfigure}
\end{figure}

\begin{figure}[t]
  \centering
  \includegraphics[width=\linewidth]{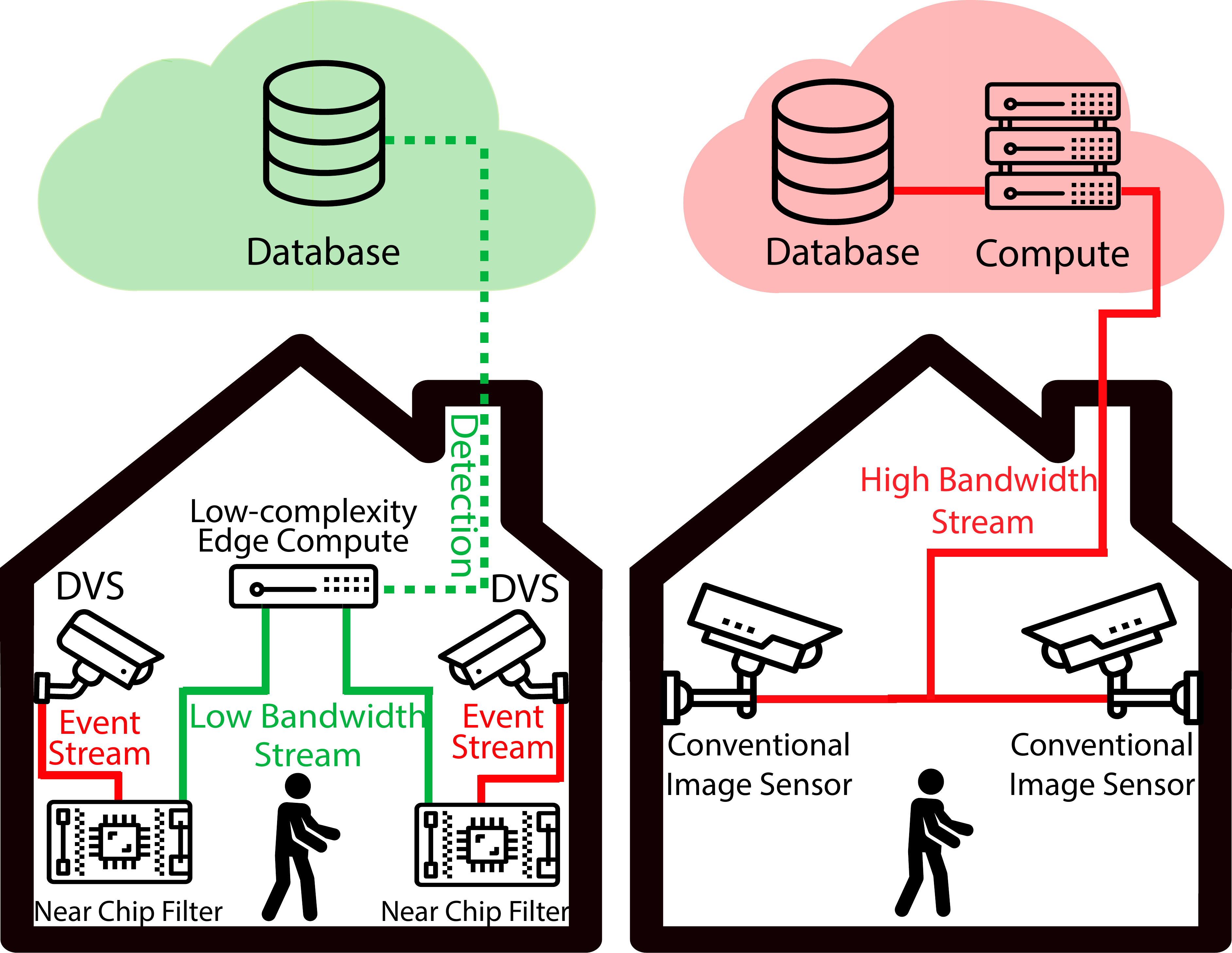}
  \caption{\gls{iot} system with \gls{dvs} for near-device
    classification. The bandwidth required to transmit the low-bandwidth
    stream is significantly lower than a compressed video stream such as
    H.265. For instance, a 10 FPS 640x480 H.265 stream may require at
    least $100\,\text{kbps}$, compared to \bandwidthtenfps{} of our
    approach, at a similar FPS.
  }
  \label{fig:dvshouse}
\end{figure}

Towards this goal, we propose a near-chip filtering architecture for pedestrian
detection (Fig.~\ref{fig:bwfigure}). Our solution
requires low bandwidth for transmitting the intermediate representations
between the sensor and the processing platform. Moreover, it
enhances privacy because of lossy subsampling, which makes it impossible
to recover the original event
representation. A single
compressed packet issued from our near-chip filter has a total length of
\framesizebits{} on average, and may be streamed through low-bitrate
channels to a
centralized networking node~(Fig. \ref{fig:dvshouse}).

\textbf{Contributions:} we have two main contributions
1)~A low-complexity hardware implementation
of an event-filter suited for
\gls{dvs} which reduces the bandwidth required to transmit the events by
up to \percentagesavings{}, targeted for a pedestrian detection system,
and 2)~An efficient detection algorithm that uses our intermediate
representation, using a 32-bit microcontroller architecture.

\section{Related Work}
\label{sec:relatedwork}

We start with an overview of related work. The use of \gls{dvs} for detection and
pattern recognition has received recent attention~\cite{Kostas2019-DVSSurvey}.
Usual target tasks include
digit recognition and simple shapes such as card
suits~\cite{timeSurfaces}, face detection~\cite{faceDetection}, and
pedestrian detection~\cite{personDetectorCars, ryu2017, Jiang2019-ICRA}. Most of
these approaches are implemented on \gls{gpu} or microprocessor-based architectures
and are not specifically targeted for \gls{iot} applications. Typically,
\gls{iot} applications require low energy consumption due to their strict energy budgets.

The asynchronous and low-bandwidth nature of event cameras
make them potentially ground-breaking for \gls{iot}.
However, \gls{dvs} sensors are
inherently noisy, making their application challenging.
Recent work addresses the filtering of \gls{dvs} noise~\cite{Khodamoradi2017-transactions, linares2017-filtering}. A
description of filtering techniques with their respective
hardware implementations is presented in~\cite{linares2019-filtersFpga}.
However, these filters are targeted for high-bandwidth applications, and
they are not specifically suited for bandwidth reduction.
Hence,
these filters are not necessarily suitable for \gls{iot} scenarios.

Several end-to-end \gls{iot} detection architectures have been proposed
as well.  In~\cite{rusci2018-thesis}, Rusci et al. showcased the advantages
of the sensor for always-on applications, by coupling an event-based
image sensor with a PULPv3 processor. While significant reductions in
energy consumption are shown in this work, the event stream is sent to
an embedded platform without any further preprocessing. Thus, this
method requires the processing platform to be near the \gls{dvs} due to
the bandwidth required to transmit the events.
In~\cite{luca2019-pcarect},
the authors present a FPGA suitable
architecture for DVS detection using \gls{pca}. While this architecture
has good performance on classification, it is not particularly targeted
for low-power architectures as they use a high-end \gls{fpga} family.
Other end-to-end
architectures, such as TrueNorth~\cite{arnon2017-cvpr-pattern} make use of
specific neuromorphic hardware to process the event stream. In this
work, the gesture recognition task is analyzed, obtaining an accuracy of
96.5\% when detecting 11 different gestures, while consuming
$200\,\text{mW}$. Compared to this work, our system has the advantage of
(i) \checkthis{a lower energy consumption}, and (ii)~\checkthis{the
capability of saving the low-bandwidth features for further
down-stream applications}.

\section{Method}

The method presented in this paper consists of two main modules.
Each one is composed of intermediate submodules.
In this section, we will describe the algorithms and present
implementation details. Specifically, we will address:
\begin{itemize}
  \item \textbf{The filtering module}: a network-aware component which runs near chip.
    It denoises the \gls{dvs} stream
    and compresses the data for further downstreaming processing. The input
    of this module is the raw event stream issued from the sensor, and the
    outputs are discrete Huffman-coded
    packets that are transmitted to the detection module.
 \item \textbf{The detection module}: receives the coded event
   representation packet
   from the filtering module, decodes it, and performs the pedestrian detection.
\end{itemize}

The combination of these two modules reduces the
filtered event bandwidth while maintaining high detection accuracy.

\begin{figure*}[t]
  \centering
  \includegraphics[width=1\linewidth]{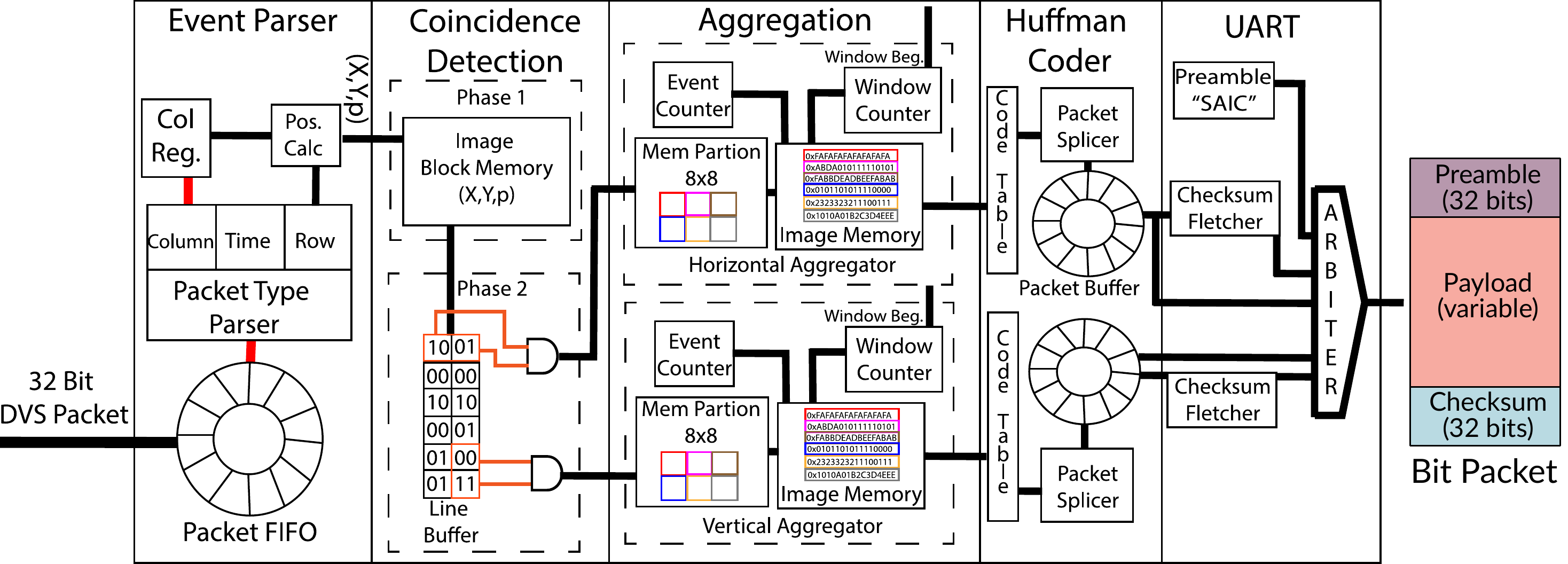}
  \caption{\gls{dvs} near-chip filter  implementation
  end-to-end diagram, displaying its submodules (Event Parser,
Coincidence Detection, Aggregation-Downsampling, Huffman Coding). For
simplicity, in this work we used an UART interface to communicate with
the detection module, but this submodule may be replaced with other communication interfaces. The packet format output fields is also displayed, showing the lengths of the different fields.}
  \label{fig:dvspipe}
\end{figure*}

\subsection{Filtering Module Description and Implementation}
The filtering module consists of four main submodules: Event Parsing,
Coincidence Detection, Aggregation-Downsampling, and Huffman
encoding.
The architecture was implemented using the
Chisel~3~\cite{bachrach2012chisel}
hardware description language.  We will describe
these submodules in this section, as well as the sensor used during our
experiments.

\subsubsection{Sensor}
We used a sensor similar to the one described
in~\cite{ryu2017-dvsarch}, with an operated resolution of $480 \times 320$ pixels.
The event rate
of our sensor was \eventrate{}. {The
DVS was connected directly to the \gls{fpga}, which was responsible for
processing the events in the \gls{gaer} packet format\cite{ryuCVPR}}.

\subsubsection{Event Parser}
\hfill

The event parser submodule translates the \gls{gaer}
representation of the sensor to a ($x$, $y$, $p$) representation, where
$x$ and $y$ are the row
and column addresses of the pixel in the sensor array, and $p$ is the polarity encoded with two
bits.
While \gls{gaer} allows for a significant bandwidth reduction between the
sensor and the \gls{fpga}, inside our architecture it is easier to work with the representation
described above for memory addressing purposes.

\textbf{Implementation:}
The Event Parsing submodule was implemented as an input \gls{fifo} queue capable of
storing 256 {G-AER} events, followed by a LUT-based decoder. The \gls{fifo} allows us to handle a rapid burst of
events from the sensor.

\subsubsection{Coincidence Detection}
\hfill

\gls{dvs} pixel arrays are susceptible to
background activity noise, which is displayed as impulse noise when \gls{dvs} events are
observed in a finite time window.
Commonly, noisy pixels will be isolated compared to signal pixels, and
thus may be removed by observing a sequence of pixel activations over space or time. Our
filter works by detecting tuples of active pixels in the vertical and
horizontal spatial directions.

The coincidence detection serves a dual purpose in our architecture: first,
it collects events in a predefined time window of length $\tau$. Then it performs a logical
AND operation between adjacent pixels. This filter is inspired by the
\gls{omd} filter described in~\cite{linares2019-filtersFpga}, but it has
two fundamental differences: (i)~we use simpler bitwise
operations between the pixels instead of a complex integrate-and-fire
model, and (ii)~a coincidence is detected only if two
pixels with the same polarity are detected.
In our architecture, $\tau =$~\windowtime{}.

\textbf{Implementation:} The coincidence detection is implemented as two
discrete memories
($M_0, M_1$) each of
size $480\times320\times2$ bits. In phase 1, $t = n\cdot\tau$, the memory array $M_0$
starts in a cleared
state, and it collects events until $t = (n + 1)\cdot\tau$, when the
window period has elapsed. In phase 2, from $t = (n+1)\cdot\tau$ until $t =
(n+2)\cdot\tau$, the
memory array $M_0$ is read and the coincidences are evaluated by observing
adjacent active vertical and horizontal pixels. At the same time, $M_1$ is collecting the
events corresponding to this time window.
The output of this submodule is composed of two channels, corresponding
to the filter applied in the vertical and horizontal dimensions. Only
active pixels are sent to the aggregation submodule.

On the \gls{fpga}, all the memory blocks were implemented with
dual-port BRAM slices. In the readout operation, a line buffer of $480$
pixels is used to store the intermediate pixels read. The coincidence
detection submodule
also propagates a signal indicating the start and end of a time window
to the aggregation submodule.

\subsubsection{Aggregation and Subsampling}
\hfill

In a static \gls{dvs} application, when binning
events in a time window, the thickness of the edge depends on
both the velocity of the object and the length of the time window. The
function of the aggregation submodule is to increase the thickness of the
edge to a normalized size before performing inference. For this, the aggregation
submodule performs successive logical OR operations across the temporal
dimension until the number of events in the aggregated frame is above a threshold. If the threshold is
not achieved in a 5$\tau$ time window, the frame buffer is
cleared and no events are propagated.


After performing the aggregation operation, an
$8\times8$ max-pooling operation is performed to the aggregated time
window.
The max-pool operation aims to reduce the scale dependency of the object
in the scene, and it reduces the dimensionality of the data.
\volkansaid{pooling reduces the scale but does not reduce the dependency
unless you are considering all scales}The
subsampling submodule operates asynchronously, only when the aggregation
submodule reports new data.

\textbf{Implementation:}
The aggregation submodule described is duplicated in order to
independently process each channel coming from the coincidence detection
submodule.
Each pixel streamed into aggregation is stored in the aggregated frame block memory ($480\times320$). At the start of every $\tau$ window, a window counter is incremented. This counter is used for implementing the temporal window integration limit of 5$\tau$.
Also, an event counter is kept for the number of pixels in the max pooled and aggregated window.
At the end of every $\tau$-sized window, the event counter is checked to be above the event threshold (1000~events). Given this condition, the aggregated frame is sent to subsampling.

The subsampling submodule is implemented using a block memory layout.
Normally to store an image in memory, a column based layout is used, where pixels are stored sequentially based on columns index.
A problem with using column indexing for max-pooling  is that for each operation different memory blocks must be accessed.
Instead, we decided to use a block memory layout:
each memory block stores pixels in the same $8\times8$ max-pooling area. Hence, a single memory read operation and comparison against 0 can perform max-pooling in a single clock cycle.
\subsubsection{Huffman encoder and Filter Interface}
\hfill

After aggregation, the output of the filter is a discrete packet of
$2 \times 60 \times 40$ bits, corresponding to the pixel readouts of
the downsampled aggregated image, for the vertical and horizontal
channel.
To further reduce the data
bandwidth, we perform a Huffman encoding using a precomputed 256-word
dictionary. On average, this results in $3.6\times$ reduction of the
payload size.

\textbf{Implementation:} The Huffman filter is implemented by storing
the codeword dictionary in BRAM and doing a lookup over the output of
the aggregation submodule.
The data is preceded by a 32-bit preamble header, and a
low-complexity Fletcher-32 checksum~\cite{fletcher1982-checksum} is
appended at the end of the packet~(Fig.~\ref{fig:dvspipe}).

For testing purposes, we streamed the event representation using an UART serial
interface between the \gls{fpga} and the detection module. Nonetheless,
other communication interfaces may be used by just changing the last
submodule in the near-chip filter. For instance, we could use the same filtering scheme with
inter-chip communication protocols, such as I2C or SPI, as well as other field-bus protocols.

\subsection{Detection Module Architecture and Implementation (BNN)}
\label{sec:detmodule}
The detection module is used to perform binary classification for
pedestrian detection from the
sparse output of the filter. It is a \gls{cnn} based architecture with
binary weights as described in~\cite{XnorNet}.

\begin{figure}[t]
  \centering

  \includegraphics[width=\linewidth]{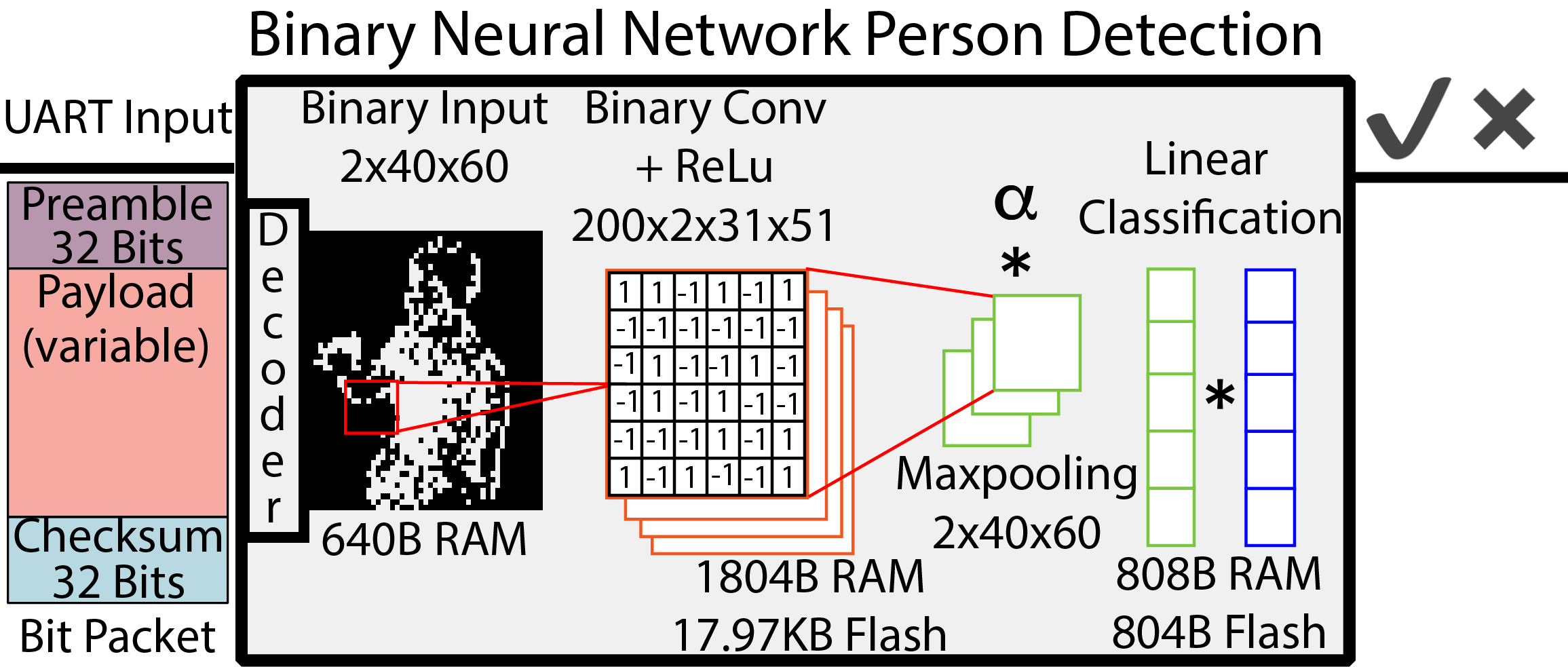}
  \caption{Binary architecture of the detector module running on the STM32F4
    microcontroller. The required memory and storage requirements are
    indicated for each submodule of the architecture.}
  \label{fig:binArch}
\end{figure}

The network architecture, presented in Fig.~\ref{fig:binArch}, is
composed of two hundred $10\times10$ convolutional filters with binary ($\pm 1$) weights.
As the output of the filter is encoded using a binary $\{0, 1\}$
representation, the convolution operation is implemented as a binary AND
operation with the positive and negative convolution filters, followed
by a population count operation.

To accelerate our calculations,
we used \gls{dsp} \gls{simd} instructions, as
well as the \gls{fpu} of the Cortex-M4 core. This
processor does not have a dedicated population count instruction
(popcnt), which is required for the neural network inference process.
Therefore, we implemented this operation as a LUT in flash. While this
approach increases the storage space required, it is a tradeoff for
faster execution speeds.

The resulting convolution value is then
scaled by a positive factor $\alpha$, followed by a ReLU nonlinearity,
and whole frame max-pooling. The detection value is obtained after a
linear readout and thresholding.

\section{Results}
\label{sec:results}
\subsection{Filtering Module Hardware Implementation Results}
The filter was synthesized on a Spartan-6
\gls{fpga}.
The maximum clock frequency achieved by our design was
{$51.18\,\text{MHz}$}.

We observed that our solution uses few resources on a low-end
\gls{fpga} platform: the
utilization of Registers is $1.10\%\,(598\,\text{slices})$,
LUTs $14.85\%\,(4053\,\text{slices})$ and DSPs
$1.72\%\,(1\,\text{slice})$. The BRAM utilization is higher,
$88.79\%\,(103\, \text{slices})$ mainly due to the
intermediate representations required to acquire events during a time
window in the coincidence detection and aggregation submodules. This utilization includes
the FIFO buffer for the \gls{dvs} packets in the event parser submodule, as
well as the Chisel~3 \texttt{DecoupledIO} interfaces used to transmit
information asynchronously between the submodules.

For reference, PCA-RECT~\cite{luca2019-pcarect} reports an utilization
of $2065\,\text{registers}$, $18238\,\text{LUTs}$, $48\,\text{BRAMs}$
and $4\,\text{DSPs}$ in a Zync-7020 SoC. Other detection architectures,
such as NullHop~\cite{FPGA-DVS-Implementation}, report an even higher
resource utilization. Our architecture
outperforms PCA-RECT on all utilization parameters but the number of
BRAMs.  This is not a surprise, as offloading the computational load of
the detection algorithm to the computing node helps keep the slice
count low.
Our approach has low utilization, yet it keeps the bandwidth
reduced without requiring a full detector implementation near-chip.

We also synthesized RetinaFilter, which is the implementation of the
background activity
filter for \gls{dvs} by Linares-Barranco et
al.~\cite{linares2019-filtersFpga}. For this, we used the same target
architecture and configuration parameters that we used in our filter.
This architecture requires $216\,\text{registers}$,
$350\,\text{LUTs}$, $16\,\text{BRAMs}$ and no DSPs. Undoubtedly, this
architecture has less slices requirements compared to our filter, but it
does not offer any gains in bandwidth reduction beyond just
denoising the image. Thus, this implementation may not be directly
applied in an \gls{iot} environment for object detection.

To assess the power consumption of the near-chip architecture, we used Xilinx
XPower Analyzer. Our module requires a total of \energyconsumption{}
at $50\,\text{MHz}$. For reference, the static power consumption
of~\cite{luca2019-pcarect} is $3\,\text{W}$ and
the dynamic power consumption is $0.37\,\text{W}$.

\subsection{Detection Module Implementation Results}

The detection module was implemented on a {STM32F429} microcontroller.
The output of the filtering module is fed into the
microcontroller using an UART port working at $115200\,\text{bps}$. Packets are
copied into the memory using \gls{dma}, and the network starts processing them
as soon as a full packet is detected and the checksum is verified.
We note that the microcontroller is kept in sleep state when
there are no packets sent through the UART, which corresponds to the
case when there is no significant activity to output events in the
\gls{dvs} event filter. This helps reduce the overall energy
consumption of the detection module.

Our network achieved an average of one inference every
\inferencetime{} with the microcontroller core running at
$180\,\text{MHz}$. It required \checkthis{$26.76\,\text{KB}$} of flash
memory to operate:
$18.76\,\text{KB}$ corresponded to our network parameters, and
$8\,\text{KB}$ were used to accelerate calculations through precomputed
look-up tables (such as the previously described population count). Finally, our network requires
\checkthis{$3.25\,\text{KB}$} of RAM to operate.


\subsection{Filtering Module Performance}
\begin{figure}
  \includegraphics[width=\linewidth]{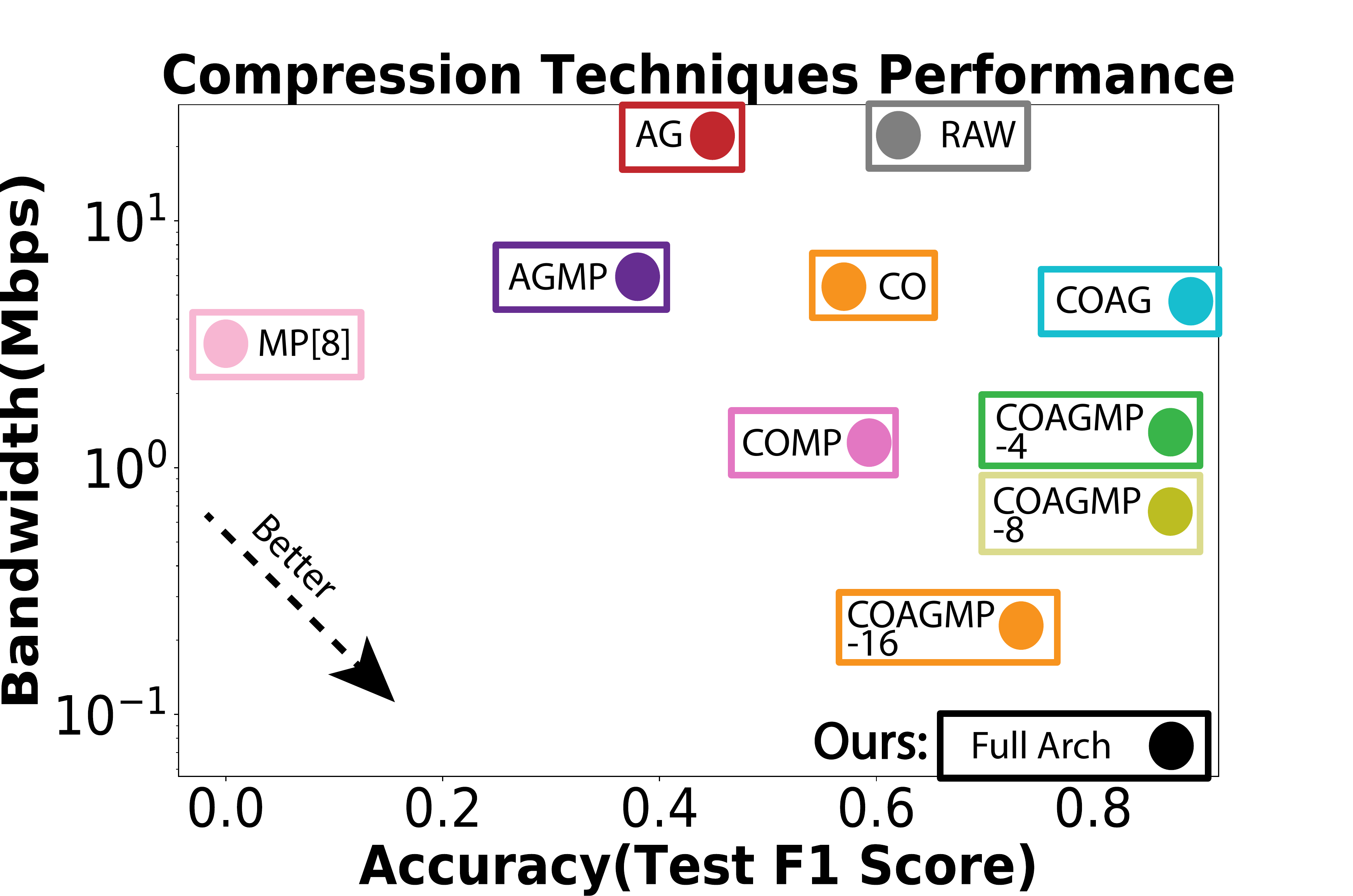}
  \caption{Testing F1 score and Bandwidth trade-off between different compression pipelines. The labels in this chart refer to: Coincidence Detection~(CO), Aggregation~(AG), Max Pooling~(MP-\#) with the max pooling ratio indicated after the dash. Our full architecture is COAGMP-8+Huffman Coding.
    The bandwidths recorded were calculated using 92380 \windowtime{} time windows. F1 score was calculated from testing on \gls{dvs} person detection dataset.
   }
  \label{fig:params}
\end{figure}

Throughout this work various approaches were tested in order to achieve
high compression with little reduction in testing accuracy. The entire
pipeline of our filtering module consists of: Coincidence Detection~(CO),
Aggregation~(AG),  Max Pooling~(MP) and Huffman Coding. Each of these
submodules reduce the bandwidth of the event stream and increase the
detection accuracy.  For the purposes of explaining  the design choices
made, we present various ablations of our method (Fig.~\ref{fig:params}). We used F1 as a metric to compare the ablations. F1 score is the harmonic mean between precision and recall scores.

To perform our measurements, we used a \gls{dvs} dataset of 273 clips of
humans, each one with a duration of 2.5s,
and 548--0.75s clips of non-humans.
The dataset was split into
a training set of 80\% and a testing set of 20\%. This dataset resulted
in 92380 \windowtime{} time windows of person and object movement. The raw
event stream bandwidth was 22.25Mb/s on average.

First, we trained our binary neural network using our full pipeline, and
we obtained a \accuracyF{} F1 testing score. Additionally, the measured bandwidth after filtering was \bandwidthfull{}.

The first ablations was removing the coincidence detection submodule. This resulted in lower testing F1 score and higher
bandwidth compared to the full pipeline. This shows the effect of the
coincidence detection removing noise: \gls{dvs} noisy events increase
bandwidth, and noisier images are harder to detect. 

The second ablation was removing the aggregation submodule. This resulted in the  testing F1 score was smaller and the output bandwidth of the filter was higher. Higher bandwidth is due to the additional frames from not temporally filtering. A lower testing F1 score without aggregation is due to less defined features for shallow learning using \glspl{bnn}.

The third ablation was changing the max-pooling size. The default value used in our pipeline was $8\times8$. When increasing this default value, bandwidth decreased and testing F1 score decreased. This is due to the lack of features due to the large dimensionality reduction.
As for decreasing the max-pooling size, bandwidth increased, yet
performance  increased by little (near 1\%). This performance increase
was small enough, that we incurred this trade off for a smaller
bandwidth model.

Our filter is capable of processing data as fast as the coincidence detection window
(\windowtime{}), resulting in the bandwidth reported below (\bandwidthfull{}).
We may further reduce the bandwidth by temporally
downsampling the detection rate, through a refractory period mechanism.
For instance, if the filter produces an output every $100\,\text{ms}$
the
final required bandwidth is \bandwidthtenfps{} on average (when some time windows are not propagated due to low event counts), and at most \bandwidthtenfpstop{}.
This enables the use of our
architecture on IoT applications using low rate wireless standards such
as 802.15.4~\cite{ieee802standard} and LoRA
and NB-IoT~\cite{sinha2017-iotspeed}.

\section{Conclusion and Further Work}

This paper introduces a novel end-to-end system for pedestrian detection
in \gls{iot} environments. Our system consists of two modules: a
near-chip filtering module and a detection module.

The near-chip filtering enabled a reduction
of up to \percentagesavings{} in bandwidth, enabling the use of \gls{dvs}
for low-bandwidth communication links. Our architecture uses few
resources on a low-end \gls{fpga}. The main bottleneck in our design was the number of BRAMs. We estimated that our module has a power
consumption of \checkthis{$91.1\,\text{mW}$}

It was shown
that, despite significant reduction in size, this representation is still useful for learning.
Additionally, it was shown that a centralized detection module may process this
representation and detect pedestrians in the scene. The computational complexity
of the detection algorithm is low because (i)~we use a shallow network
with a low number of feature detectors, and (ii)~the use of a binary
network representation reduces the execution time on a low-end
microcontroller.

Some future work for this investigation involves implementing
applications fully on the edge. This would require integrating the
filtering algorithm along with the detection algorithm. Therefore, one would benefit from the
additional bandwidth savings of only sending the detection
result over the wire rather than a representation. Due to the optimized
nature of our filter for a single detection task, we expect to get
better results compared to~\cite{luca2019-pcarect}.

Another direction for future work would be an on-chip \gls{asic}
implementation combining our filtering algorithm with the \gls{dvs}. This
would produce additional power savings along with bandwidths savings, by
removing the need to go off-sensor for filtering.

Finally, an interesting approach for this work would be to perform classification based on multiple \gls{dvs} streams from different cameras. As the output of the filter is lightweight, we could imagine using multiple sensors for a single classification performed on the low complexity edge compute.
\fixme{Ref needed}

\bibliographystyle{IEEEbib}
\bibliography{isvlsi_bib}

\begin{thebibliography}{10}

\bibitem{edgeFacebook}
Carole-Jean Wu, David Brooks, Kevin Chen, Douglas Chen, Sy~Choudhury, Marat
  Dukhan, Kim Hazelwood, Eldad Isaac, Yangqing Jia, Bill Jia, et~al.,
\newblock ``Machine learning at facebook: Understanding inference at the
  edge,''
\newblock in {\em 2019 IEEE International Symposium on High Performance
  Computer Architecture (HPCA)}. IEEE, 2019, pp. 331--344.

\bibitem{Kostas2019-DVSSurvey}
Guillermo Gallego, Tobi Delbruck, Garrick Orchard, Chiara Bartolozzi, Brian
  Taba, Andrea Censi, Stefan Leutenegger, Andrew Davison, Joerg Conradt, Kostas
  Daniilidis, et~al.,
\newblock ``{Event-based vision: A survey},''
\newblock {\em arXiv:1904.08405}, 2019.

\bibitem{timeSurfaces}
Xavier Lagorce, Garrick Orchard, Francesco Galluppi, Bertram~E Shi, and Ryad~B
  Benosman,
\newblock ``{Hots: a hierarchy of event-based time-surfaces for pattern
  recognition},''
\newblock {\em IEEE transactions on pattern analysis and machine intelligence},
  vol. 39, no. 7, pp. 1346--1359, 2016.

\bibitem{faceDetection}
Souptik Barua, Yoshitaka Miyatani, and Ashok Veeraraghavan,
\newblock ``{Direct face detection and video reconstruction from event
  cameras},''
\newblock in {\em 2016 IEEE WACV}. IEEE, 2016, pp. 1--9.

\bibitem{personDetectorCars}
Rohan Ghosh, Abhishek Mishra, Garrick Orchard, and Nitish~V Thakor,
\newblock ``{Real-time object recognition and orientation estimation using an
  event-based camera and CNN},''
\newblock in {\em 2014 IEEE BioCAS}, 2014, pp. 544--547.

\bibitem{ryu2017}
Jia Li, Feng Shi, Wei-Heng Liu, Dongqing Zou, Qiang Wang, Paul~KJ Park, and
  Hyunsurk Ryu,
\newblock ``Adaptive temporal pooling for object detection using dynamic vision
  sensor.,''
\newblock in {\em BMVC}, 2017.

\bibitem{Jiang2019-ICRA}
Z.~{Jiang}, P.~{Xia}, K.~{Huang}, W.~{Stechele}, G.~{Chen}, Z.~{Bing}, and
  A.~{Knoll},
\newblock ``Mixed frame-/event-driven fast pedestrian detection,''
\newblock in {\em 2019 International Conference on Robotics and Automation
  (ICRA)}, May 2019, pp. 8332--8338.

\bibitem{Khodamoradi2017-transactions}
A.~{Khodamoradi} and R.~{Kastner},
\newblock ``O(n)-space spatiotemporal filter for reducing noise in neuromorphic
  vision sensors,''
\newblock {\em IEEE Transactions on Emerging Topics in Computing}, pp. 1--1,
  2017.

\bibitem{linares2017-filtering}
A.~{Linares-Barranco}, F.~{Gómez-Rodríguez}, V.~{Villanueva},
  L.~{Longinotti}, and T.~{Delbrück},
\newblock ``{A USB3.0 FPGA event-based filtering and tracking framework for
  dynamic vision sensors},''
\newblock in {\em 2015 IEEE International Symposium on Circuits and Systems
  (ISCAS)}, May 2015, pp. 2417--2420.

\bibitem{linares2019-filtersFpga}
A.~{Linares-Barranco}, F.~{Perez-Peña}, D.~P. {Moeys}, F.~{Gomez-Rodriguez},
  G.~{Jimenez-Moreno}, S.~{Liu}, and T.~{Delbruck},
\newblock ``Low latency event-based filtering and feature extraction for
  dynamic vision sensors in real-time fpga applications,''
\newblock {\em IEEE Access}, vol. 7, pp. 134926--134942, 2019.

\bibitem{rusci2018-thesis}
Manuele Rusci, Davide Rossi, Eric Flamand, Massimo Gottardi, Elisabetta
  Farella, and Luca Benini,
\newblock ``{Always-ON} visual node with a hardware-software event-based
  binarized neural network inference engine,''
\newblock in {\em Proceedings of the 15th ACM International Conference on
  Computing Frontiers}. ACM, 2018, pp. 314--319.

\bibitem{luca2019-pcarect}
Bharath Ramesh, Andr{\'{e}}s Ussa, Luca~Della Vedova, Hong Yang, and Garrick
  Orchard,
\newblock ``{PCA-RECT:} an energy-efficient object detection approach for event
  cameras,''
\newblock {\em CoRR}, vol. abs/1904.12665, 2019.

\bibitem{arnon2017-cvpr-pattern}
A.~{Amir}, B.~{Taba}, D.~{Berg}, T.~{Melano}, J.~{McKinstry}, C.~D. {Nolfo},
  T.~{Nayak}, A.~{Andreopoulos}, G.~{Garreau}, M.~{Mendoza}, J.~{Kusnitz},
  M.~{Debole}, S.~{Esser}, T.~{Delbruck}, M.~{Flickner}, and D.~{Modha},
\newblock ``A low power, fully event-based gesture recognition system,''
\newblock in {\em 2017 IEEE Conference on Computer Vision and Pattern
  Recognition (CVPR)}, July 2017, pp. 7388--7397.

\bibitem{bachrach2012chisel}
Jonathan Bachrach, Huy Vo, Brian Richards, Yunsup Lee, Andrew Waterman, Rimas
  Avi{\v{z}}ienis, John Wawrzynek, and Krste Asanovi{\'c},
\newblock ``Chisel: constructing hardware in a scala embedded language,''
\newblock in {\em DAC Design Automation Conference 2012}. IEEE, 2012, pp.
  1212--1221.

\bibitem{ryu2017-dvsarch}
Bongki Son, Yunjae Suh, Sungho Kim, Heejae Jung, Jun-Seok Kim, Changwoo Shin,
  Keunju Park, Kyoobin Lee, Jinman Park, Jooyeon Woo, et~al.,
\newblock ``{4.1 A 640$\times$ 480 dynamic vision sensor with a 9$\mu$m pixel
  and 300Meps address-event representation},''
\newblock in {\em 2017 IEEE ISSCC}, 2017, pp. 66--67.

\bibitem{ryuCVPR}
Yoel Yaffe, Nathan Levy, Evgeny Soloveichik, Sebastien Derhy, Ayal Keisar, Elad
  Rozin, Liron Artsi, Jun-Seok Kim, Keunju Park, Bongki Son, Yunjae Suh, Heejae
  Jung, Changwoo Shin, Jooyeon Woo, Yohan Roh, Hyunku Lee, and Hyunsurk(Eric)
  Ryu,
\newblock ``Dynamic vision sensor: The road to market,''
  \url{http://rpg.ifi.uzh.ch/docs/ICRA17workshop/Samsung.pdf}.

\bibitem{fletcher1982-checksum}
J.~{Fletcher},
\newblock ``An arithmetic checksum for serial transmissions,''
\newblock {\em IEEE Transactions on Communications}, vol. 30, no. 1, pp.
  247--252, January 1982.

\bibitem{XnorNet}
Mohammad Rastegari, Vicente Ordonez, Joseph Redmon, and Ali Farhadi,
\newblock ``{XNOR-Net}: {ImageNet} classification using binary convolutional
  neural networks,''
\newblock {\em arXiv:1603.05279}, 2016.

\bibitem{FPGA-DVS-Implementation}
Alejandro Linares-Barranco, Antonio Rios-Navarro, Ricardo Tapiador-Morales, and
  Tobi Delbruck,
\newblock ``{Dynamic Vision Sensor integration on FPGA-based CNN accelerators
  for high-speed visual classification},''
\newblock {\em arXiv:1905.07419}, 2019.

\bibitem{ieee802standard}
{C/LM - LAN/MAN Standards Committee},
\newblock ``{802.15.4-2015 - IEEE Standard for Low-Rate Wireless Networks},''
  {https://standards.ieee.org/content/ieee-standards/en/standard/802\_15\_4-2015.html},
  2015.

\bibitem{sinha2017-iotspeed}
Rashmi~Sharan Sinha, Yiqiao Wei, and Seung-Hoon Hwang,
\newblock ``A survey on lpwa technology: Lora and nb-iot,''
\newblock {\em Ict Express}, vol. 3, no. 1, pp. 14--21, 2017.

\end{thebibliography}

\end{document}